\documentclass[runningheads]{llncs}

\usepackage[T1]{fontenc}
\usepackage{microtype}
\usepackage{graphicx}
\usepackage{booktabs}
\usepackage{amsmath}
\usepackage{amssymb}
\usepackage{multirow}
\usepackage{xcolor}
\PassOptionsToPackage{hyphens}{url}
\usepackage{url}
\usepackage{hyperref}
\hypersetup{hidelinks}

\graphicspath{{figures/}}

\AddToHookNext{shipout/foreground}{%
  \put(36,-28){%
    \makebox(0,0)[lt]{%
      \normalfont\fontsize{8}{10}\selectfont
      \begin{tabular}{@{}l@{}}
        Accepted by the 33rd International Conference on Neural Information Processing (ICONIP 2026).\\
        Submitted: March 17, 2026.\quad Accepted: August 14, 2026.\\
        Conference: November 23--27, 2026, Melbourne, Australia.
      \end{tabular}%
    }%
  }%
}

\begin{document}

\title{When Self-Consistency Hurts: The Hidden Cost\\
of Majority Voting in Small Language Models}
\titlerunning{When Self-Consistency Hurts}

\author{%
Wanrong Yang$^{\dagger}$\inst{1}\and
Lingfang Li$^{\dagger}$\inst{1}\and
Jun Zhang\inst{2}\and
Dominik Wojtczak\inst{1}}
\authorrunning{W.~Yang et al.}
\institute{%
Department of Computer Science, University of Liverpool, Liverpool, United Kingdom\\
\email{\{Wanrong.Yang,L.Li85,D.Wojtczak\}@liverpool.ac.uk}
\and
LIPEX, Shenzhen, China\\
\email{zhangjunlipex@gmail.com}}

\maketitle
\renewcommand{\thefootnote}{}
\footnotetext{$^{\dagger}$\,These authors contributed equally to this work.}
\renewcommand{\thefootnote}{\arabic{footnote}}

\begin{abstract}
We study prompted chain-of-thought (CoT) with self-consistency
(SC), i.e.\ sampling $N$ traces and plurality-voting, on small
language models, a regime where per-query compute budgets are
tight yet SC is commonly applied with a fixed sample budget
inherited from large-model practice, regardless of instance
difficulty. Across two model
families (Qwen-2.5-0.5B, Gemma-3-1B) at FP16 and MLX-Q4 (4-bit quantisation in Apple's MLX framework), on GSM8K
and MATH-500, we run 8 per-task and 4 task-pooled experiments at
$n{=}200$--$400$ with $1\,000{\times}$ paired-bootstrap confidence
intervals (CIs). Our first result is an upper bound: an oracle
gate deciding per prompt whether to invoke SC reaches
$109$--$174\%$ of always-SC's accuracy at $23$--$30\%$ of its
compute in all $12/12$ cases, so at $N{=}5$ always-SC is dominated
everywhere on our grid. That margin is exactly the rate at which
SC flips \emph{correct greedy answers to wrong}, here $3$--$10.5\%$
of prompts, which on Qwen-Q4${\times}$MATH-500 exceeds the
wrong-to-right rate and leaves always-SC $3$\,pt \emph{worse} than
always-greedy. Our second result is a deployable approximation: a
gate over greedy-pass signals skips $45$--$99\%$ of SC calls, with
bootstrap CI half-widths under $5$\,pt, removing $36$--$79\%$ of
always-SC compute. Accuracy retention at those operating points has point estimates of
$98.1$--$98.8\%$, which we report as point estimates because their
CIs are wide (lower bounds $76$--$86\%$) over an SC accuracy of
only $15$--$36\%$. The evidence covers $N{=}5$ on two ${\le}1$B
families and two math benchmarks, which is the scope we claim.
Code is available at \url{https://github.com/heywanrong/SiftSC}.

\end{abstract}

\section{Introduction}
\label{sec:intro}

Test-time compute (TTC) has become a standard component of modern
language-model deployment. Starting from chain-of-thought (CoT)
prompting~\cite{Wei2022CoT,Kojima2022ZeroShotCoT}, a family of
methods spanning self-consistency (SC) sampling~\cite{Wang2023SC},
best-of-$N$ reranking with learned
verifiers~\cite{Cobbe2021Verifiers,Lightman2023LetsVerify},
tree-structured search~\cite{Yao2023ToT}, budget-forced
CoT~\cite{Snell2024TTC} and tool-integrated
self-verification~\cite{Kang2025T1} now underpins the strongest
open and proprietary
systems~\cite{OpenAIO1,Chen2025LongCoTSurvey}. The
prevailing deployment recipe is simple: for every query, sample
$N$ traces and plurality-vote. This \emph{always-SC} policy is the
default in every widely-used inference stack we are aware of, and
it has been validated extensively on large ($\ge 7$B) models, where
$N{=}5$ to $N{=}40$ reliably trades compute for
accuracy~\cite{Chen2025LongCoTSurvey,Snell2024TTC}.

The economics look different for \textbf{small} language models,
the segment where inference cost matters most: on-device
deployment is bounded by memory and
battery~\cite{Wang2025MNNLLM,Bsharat2025MobileMMLU}, cloud cost is
dominated by latency~\cite{Cao2026PER}, privacy pressures favour
local inference, and distilled sub-4B models now match or beat
$7$B$+$ reasoners on math~\cite{Xu2025Phi4Mini}. The always-SC recipe was nonetheless
inherited from large-model practice without much scrutiny of
whether it remains the right default for \emph{prompted} CoT at
this scale. A parallel literature on efficient TTC has meanwhile
developed around cutting the sample
count~\cite{Aggarwal2023AdaptiveConsistency,Wan2024RASC}, adding
confidence-aware stopping~\cite{Aghazadeh2025CGES,Kim2026ReASC},
and predicting the budget up front~\cite{Ji2025SeerSC}. All of it
assumes that SC at some $N{>}1$ is worth running. On the models we
test, that premise is what fails.

We revisit two questions. \textbf{(Q1)~Is always-SC actually
worth it on small models?} The commonly used metric is SC's
accuracy relative to greedy, but the realistic deployment baseline
is not plain greedy: it is ``run cheap greedy first, and run SC
only when the prompt requires it''. The fair comparison for always-SC is therefore an \emph{oracle} gate that picks the better
of greedy and SC per prompt. \textbf{(Q2)~If always-SC is
dominated, can a cheap gate approximate that oracle?} If
so, \emph{selective} SC becomes the natural default at this scale.

We answer both concretely. On
Qwen-2.5-0.5B-Instruct~\cite{Qwen25Report} and
Gemma-3-1B-it~\cite{Gemma3Report}, each in FP16 and 4-bit
quantisation in Apple's MLX framework, across GSM8K~\cite{Cobbe2021Verifiers} and
MATH-500~\cite{Hendrycks2021MATH}, the oracle gate dominates
always-SC in all $12/12$ experiments, retaining $109$--$174\%$ of
its accuracy for $23$--$30\%$ of its compute. What drives the
dominance is a non-trivial \emph{harm rate}: SC flips correct
greedy answers to wrong on $3$--$10.5\%$ of prompts, comparable to
and sometimes larger than the rate at which it flips wrong to
right. On Qwen-Q4${\times}$MATH-500 the harm rate exceeds the help rate, and always-SC ends up $3$\,pt worse than
always-greedy. A gate on greedy-pass signals recovers much of this margin. On the four pooled cells it skips $45$--$99\%$ of SC
calls, with $1\,000{\times}$ bootstrap confidence intervals (CIs)
whose half-widths on skip rate stay under $5$\,pt, and removes $36$--$79\%$ of
always-SC compute. Retention of always-SC accuracy at those
operating points is $98.1$--$98.8\%$ as a point estimate, and we report it only as such, because its bootstrap CIs are wide
(\S\ref{sec:pooled}).

\paragraph{Scope.}
Our evidence is narrower than the framing above may suggest, so we state the boundary here. Everything below is at $N{=}5$, on two
${\le}1$B families, at two precisions, on two math benchmarks, with
one generation seed per cell. Within that grid the dominance of the
per-prompt oracle gate over always-SC holds in every cell and the harm rate is
measurable. We do not claim either for larger models,
reasoning-distilled models, non-mathematical tasks, or larger $N$
(\S\ref{sec:limitations}).

\paragraph{Relation to prior efficient-TTC work.}
Adaptive-SC methods~\cite{Aggarwal2023AdaptiveConsistency,Wan2024RASC,Aghazadeh2025CGES,Kim2026ReASC,Ji2025SeerSC,Liu2026PETS}
reduce $N$ once SC has been invoked, whereas our gate acts on the
invocation decision itself. Cascades and
routers~\cite{Zheng2025CITER,She2025TokenRouter,Ong2024RouteLLM,Chen2024FrugalGPT,Chuang2024ConfidenceRoute}
choose \emph{which} model answers, while our gate stays inside one
model. \S\ref{sec:related} compares it with both lines, and
\S\ref{sec:limitations} notes that we run no head-to-head comparison with either.

\paragraph{Contributions.}
(i)~The first per-prompt measurement of prompted SC's
``helps vs.\ hurts'' decomposition on small language models,
exhibiting a non-negligible, and at one cell \emph{dominant}, harm rate (\S\ref{sec:oracle}); (ii)~evidence against the ``SC is
free'' premise of the efficient-TTC literature, bounded to $N{=}5$
on these two families and two math benchmarks; (iii)~a minimal
1--12 feature cross-validated TTC-gate that captures much of the
oracle's savings with tight skip-rate CIs, plus an ablation in
which added capacity (a multi-layer perceptron, MLP) brings no benefit at this data scale (\S\ref{sec:method}); (iv)~a
single-laptop reproduction path under MLX, included in the protocol.

\section{Background and Related Work}
\label{sec:related}

\paragraph{Test-time compute.}
CoT prompting~\cite{Wei2022CoT,Kojima2022ZeroShotCoT}
and SC~\cite{Wang2023SC} established the
sample-and-aggregate recipe, \cite{Snell2024TTC} formalised its
compute--accuracy scaling for large models, and
\cite{Chen2025LongCoTSurvey} surveys the long-CoT literature that
followed. \cite{Yao2023ToT} extends the recipe to structured
search;
\cite{Lightman2023LetsVerify,Cobbe2021Verifiers} train process-
and outcome-reward verifiers for reranking. None of this line
questions the assumption that SC is worth running on every
prompt.

\paragraph{Adaptive self-consistency.}
Three families of methods make SC cheaper. Early-stopping rules
halt sampling once the running majority looks stable, whether on
count-based convergence~\cite{Aggarwal2023AdaptiveConsistency},
Bayesian and reliability-aware
criteria~\cite{Aghazadeh2025CGES,Kim2026ReASC}, or votes
reweighted by per-sample reasoning quality~\cite{Wan2024RASC}.
Up-front budget prediction fixes $N$ before any sampling, as in
Seer's System-1 entropy estimate~\cite{Ji2025SeerSC}. Cross-prompt
allocation spends one global budget where it helps the most~\cite{Liu2026PETS}. A fourth line provides statistical guarantees for sampling-based
TTC~\cite{Zhou2025RPC,Cordero2025Certified}. All of them take the
invocation of SC as given and choose how many traces to draw,
which is why the gate composes with them rather than competing with them: it
answers the prior question of whether to draw any. The difference
is quantitative too, since an early-stopping rule cutting $N$ from
$5$ to $2$ saves at most $2.5\times$ on a prompt, while skipping
SC saves the whole SC budget. We do not run any of them
head-to-head against the gate (\S\ref{sec:limitations}).

The closest prior observation that more sampling can hurt is
\cite{Chen2024Escape}, who show majority-vote compound systems to
be non-monotone in the number of model calls and trace this to
heterogeneous query difficulty, easy queries improving while hard
ones degrade. Their analysis picks one $N$ per task on large
models. We measure the per-prompt help/harm decomposition on small
models and act on it per prompt.

\paragraph{Small language models and quantisation.}
Distilled small language
models~\cite{Kang2025T1,Xu2025Phi4Mini} make the
small-language-model ${+}$ TTC setting practically relevant. \cite{Li2025QuantMath} report that
4-bit post-training quantisation (e.g.\ AWQ~\cite{Lin2024AWQ},
GPTQ~\cite{Frantar2023GPTQ}) costs Llama-3 models an average of
$11\%$ and up to $32\%$ of their math-reasoning accuracy. We measure a $5.5$--$27$\,pt FP16-vs-Q4 drop on
our two families and show it interacting adversely with SC on at
least one cell.

\paragraph{Calibration and verbalised confidence.}
Classical calibration work~\cite{Guo2017Calibration} and
large-language-model uncertainty
studies~\cite{Xiong2023Uncertainty,Chuang2024ConfidenceRoute} show
token-level confidences to be useful though imperfect signals. We use
only native greedy-pass logits and simple text features, with no
fine-tuning and no verbaliser.

\paragraph{Routing and cascades.}
Cascade and router
systems~\cite{Zheng2025CITER,She2025TokenRouter,Ong2024RouteLLM,Chen2024FrugalGPT}
pick between models of different sizes or providers from
logit-level or learned signals. Our gate is \emph{within-model}
and decides not \emph{which} model runs but \emph{whether} to run
TTC at all, so it needs no cloud teacher, external verifier, or
auxiliary model.

\section{Experimental Setup}
\label{sec:setup}

\paragraph{Models.}
We use Qwen-2.5-0.5B-Instruct~\cite{Qwen25Report} and
Gemma-3-1B-it~\cite{Gemma3Report}, each at two precisions: MLX-FP16
(native BF16 cast to FP16) and MLX-Q4 (affine weight-only
quantisation, group size~$64$). On Apple Silicon the standard
bitsandbytes-NF4 (4-bit NormalFloat) and auto-gptq stacks either lack GPU kernels or
fall back to CPU at $\ge 10\times$ latency, leaving MLX-Q4 as the
practical Apple-GPU path. The combination of two families and two precisions gives four cells, each run on both datasets. A cell on one dataset is a per-task cell; a cell on both datasets is a pooled cell.

\paragraph{Datasets.}
We use the GSM8K test
split~\cite{Cobbe2021Verifiers} and MATH-500, the HuggingFaceH4
$500$-problem evaluation subset of
MATH~\cite{Hendrycks2021MATH,Lightman2023LetsVerify}. Each cell
runs the first $200$ problems per dataset, giving $1\,600$
prompt-level observations across the per-task grid, and we pool the
two tasks into a $400$-per-cell aggregate to test whether the gate
transfers across tasks.

\paragraph{Prompt and sampling protocol.}
One in-context exemplar is followed by the zero-shot CoT
cue~\cite{Kojima2022ZeroShotCoT} ``Let's think step by step.''
Sampled traces use temperature $0.7$ and top-$p$ $0.95$; the
deterministic pass uses $T{=}0$. Per prompt we collect one greedy
answer and $N{=}5$ sampled answers, and SC@$5$ is plurality voting
over the $N$ samples with a sum-log-probability tiebreak. Answer
parsing is a layered regex over \verb|\boxed{}|, ``The answer is
$X$'', ``\verb|####|'', and finally the last numeric expression in
the output. All runs use seed~$1$ and deterministic MLX sampling.

\paragraph{Evaluation.}
Raw accuracies are reported as point estimates.
FP16-vs-Q4 deltas use paired $1\,000{\times}$
bootstrap at the prompt level. Gate operating points at fixed
threshold $\tau$ use per-cell $1\,000{\times}$ bootstrap over
prompts. Where we report \emph{retention} $R(\tau) =
\text{acc}_\text{gate}(\tau) / \text{acc}_\text{always-SC}$, the
denominator is the empirical SC@$5$ accuracy on the same cell.
Two properties of these intervals matter. Every bootstrap
resamples prompts while holding each prompt's greedy and SC outcomes fixed, so an interval says how much a conclusion depends
on which prompts were drawn, not how much it would move under a
different generation seed. With one seed per cell, sampling
variance in SC itself is not covered by any interval we report.
Retention is also a ratio with a small denominator here (SC
accuracy is $15$--$36\%$), so its intervals are wide even where
the point estimate is stable. We therefore keep skip rate and
retention as separate claims of different strengths throughout.

\paragraph{Cost accounting.}
We count compute in units of one generation pass, which keeps the
accounting hardware-independent at the cost of ignoring
prompt-length, batching and memory effects. Always-greedy costs
$1$ and always-SC costs $N$. A gate always pays for the greedy
pass, which provides both the fallback answer and the gate
features, so a skipped prompt costs exactly $1$ pass and nothing
more. When the gate invokes SC we charge $N$, treating the greedy
trace as one of the $N$ votes and drawing $N{-}1$ further samples:
\begin{equation}
C \;=\; 1 + (N-1)\,p_{\mathrm{invoke}},
\label{eq:cost}
\end{equation}
where $p_{\mathrm{invoke}}$ is the fraction of prompts on which
the gate invokes SC. Our measured SC@$5$ accuracies come from five
temperature-sampled traces rather than four plus the greedy one,
so Eq.~\ref{eq:cost} is mildly idealised: an implementation requiring five fresh samples pays $1{+}N$ on invoked prompts,
and the compute removed from always-SC is then $25$--$79\%$
rather than the $36$--$79\%$ we report.

\paragraph{Hardware and compute budget.}
Everything runs on one MacBook Pro M4 Pro ($24$\,GB unified
memory, $16$-core GPU, MLX 0.31, \texttt{mlx-lm} 0.31, PyTorch
2.11 with Metal Performance Shaders (MPS), Python 3.13), with no CUDA and no cloud. Generation
takes ${\sim}6.7$ GPU-hours across the four cells, result consolidation under three minutes, and the released code, cached records and
seeds reproduce every table and figure end-to-end.

\section{The Oracle-Dominance Finding}
\label{sec:oracle}

\subsection{Structural Decomposition}

For any prompt $x\in\mathcal{X}$, let $g(x),s(x)\in\{0,1\}$
indicate whether greedy and SC@$5$ are correct, and define the
\emph{help} and \emph{harm} indicators $y(x)=\mathbf{1}[s(x){=}1
\wedge g(x){=}0]$ and $h(x)=\mathbf{1}[g(x){=}1 \wedge
s(x){=}0]$. Writing $\pi$ for the test distribution and
$\mathbb{E}$ for $\mathbb{E}_\pi$, the accuracies of greedy,
always-SC, and a per-prompt oracle that picks the better of the
two are
\begin{align}
\alpha_\mathrm{g}   &\;=\; \mathbb{E}[g], &
\alpha_\mathrm{sc}  &\;=\; \mathbb{E}[g] + \mathbb{E}[y] - \mathbb{E}[h], &
\alpha_\mathrm{o}   &\;=\; \mathbb{E}[g] + \mathbb{E}[y],
\end{align}
where $\alpha_\mathrm{sc}=\mathbb{E}[s]$ and
$\alpha_\mathrm{o}=\mathbb{E}[\max(g,s)]$. We write
$\Delta_\mathrm{sc}=\mathbb{E}[y]-\mathbb{E}[h]=\alpha_\mathrm{sc}-\alpha_\mathrm{g}$
for SC's net gain over greedy. The oracle's gain over always-SC is
then exactly
\begin{equation}
\alpha_\mathrm{o} - \alpha_\mathrm{sc} \;=\; \mathbb{E}[h],
\label{eq:oracle-gap}
\end{equation}
so \emph{always-SC is strictly dominated whenever}
$\mathbb{E}[h]>0$. The community's view of SC implicitly assumes $\mathbb{E}[h]\ll\mathbb{E}[y]$, which holds for large
models but, as we show below, not for small ones.

\subsection{Empirical Measurement}

\begin{figure}[tb]
\centering
\includegraphics[width=0.60\linewidth]{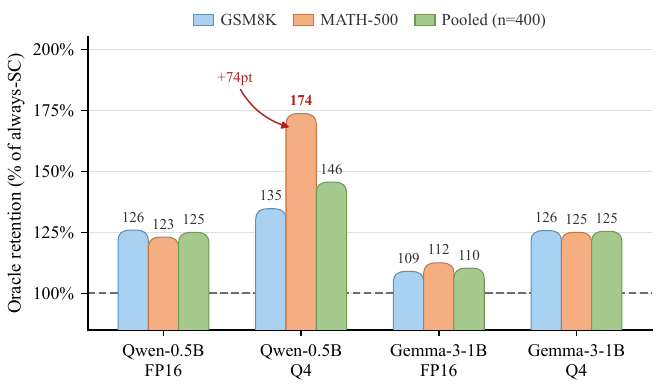}
\caption{Oracle TTC-gate retention
$\rho=\alpha_\mathrm{o}/\alpha_\mathrm{sc}$ across all 12
experiments (4 model-precision cells $\times$ 2 datasets $+$
pooled). The annotated Qwen-0.5B${\times}$Q4${\times}$MATH-500
bar is the worst cell, where the oracle is $7$\,pt more accurate than always-SC, a retention of $174\%$. The dashed line at $100\%$ marks always-SC parity, and
retention lies in $[109\%,174\%]$ everywhere.}
\label{fig:oracle}
\end{figure}

Figure~\ref{fig:oracle} plots the oracle-retention ratio
$\rho=\alpha_\mathrm{o}/\alpha_\mathrm{sc}$ across all $12$
experiments ($4$ model-precision cells $\times$ ($2$ datasets $+$
pooled)). Every bar lies at or above the $100\%$ always-SC
reference, with the $174\%$ outlier at Qwen-0.5B-Q4 on MATH-500,
where always-SC leaves $7$\,pt of reachable accuracy unused. The
rest fall between $109\%$ (Gemma-FP16 on GSM8K) and $146\%$
(Qwen-Q4 pooled), and pooled retention ($110$--$146\%$) tracks
per-task retention, as is desirable for a gate meant to
transfer across tasks. The oracle runs SC only on prompts where
SC helps, so by Eq.~\ref{eq:cost} its compute is
$1+(N-1)\,\mathbb{E}[y]$. With the help rates of
Table~\ref{tab:yh}, this is $23$--$30\%$ of always-SC's cost in
all $12$ experiments.

Table~\ref{tab:yh} lists the underlying accuracies. FP16 greedy
accuracies span $18$--$39\%$ and Q4 greedy accuracies
$10$--$20.5\%$. The paired FP16-vs-Q4 drop runs from $5.5$ to
$27$\,pt across greedy and SC@$5$, and all eight $95\%$
paired-bootstrap CIs exclude $0$ (half-widths $4.8$--$7.5$\,pt). The SC@$5$-minus-greedy gain is $+1.5$ to
$+7$\,pt on seven of the eight per-task cells and $-3$\,pt on
Qwen-Q4${\times}$MATH-500. Even the weakest claim for SC,
that it is never worse than greedy, fails at this scale.

\begin{table}[tb]
\centering\small
\caption{Per-cell accuracies and the per-prompt decomposition of SC's outcomes, $n{=}200$ per row. $\alpha_\mathrm{g}$ and
$\alpha_\mathrm{sc}$ are greedy and SC@$5$ accuracy,
$\mathbb{E}[y]$ the ``SC helps'' rate, $\mathbb{E}[h]$ the ``SC
hurts'' rate, and
$\Delta_\mathrm{sc}=\mathbb{E}[y]-\mathbb{E}[h]$ the net
SC-minus-greedy gain (pt). On Qwen-Q4${\times}$MATH-500,
$\mathbb{E}[h]>\mathbb{E}[y]$ and SC is net-negative.}
\label{tab:yh}
\setlength{\tabcolsep}{5pt}
\begin{tabular}{@{}llcccccr@{}}
\toprule
Cell & Dataset & $\alpha_\mathrm{g}$ & $\alpha_\mathrm{sc}$ & $\mathbb{E}[y]$ & $\mathbb{E}[h]$ & $\Delta_\mathrm{sc}$ \\
\midrule
Qwen-FP16  & GSM8K    & $39.0\%$ & $40.5\%$ & $12.0\%$ & $10.5\%$ & $+1.5$ \\
Qwen-Q4    & GSM8K    & $20.5\%$ & $24.5\%$ & $12.5\%$ &  $8.5\%$ & $+4.0$ \\
Gemma-FP16 & GSM8K    & $37.5\%$ & $44.5\%$ & $11.0\%$ &  $4.0\%$ & $+7.0$ \\
Gemma-Q4   & GSM8K    & $15.0\%$ & $17.5\%$ &  $7.0\%$ &  $4.5\%$ & $+2.5$ \\
Qwen-FP16  & MATH-500 & $18.0\%$ & $19.5\%$ &  $6.0\%$ &  $4.5\%$ & $+1.5$ \\
\textbf{Qwen-Q4} & \textbf{MATH-500} & $12.5\%$ & $9.5\%$ &  $4.0\%$ &  $7.0\%$ & $\mathbf{-3.0}$ \\
Gemma-FP16 & MATH-500 & $22.5\%$ & $28.0\%$ &  $9.0\%$ &  $3.5\%$ & $+5.5$ \\
Gemma-Q4   & MATH-500 & $10.0\%$ & $12.0\%$ &  $5.0\%$ &  $3.0\%$ & $+2.0$ \\
\bottomrule
\end{tabular}
\end{table}

The same table gives the prompt-level outcome decomposition. The
help rate $\mathbb{E}[y]$ runs from $4\%$ (Qwen-Q4 on MATH-500) to
$12.5\%$ (Qwen-Q4 on GSM8K) and the harm rate $\mathbb{E}[h]$ from
$3\%$ (Gemma-Q4 on MATH-500) to $10.5\%$ (Qwen-FP16 on GSM8K), so harm
is non-trivial in every cell, and on Qwen-Q4${\times}$MATH-500 its
$7\%$ against $4\%$ of help inverts the sign of
$\Delta_\mathrm{sc}$. The oracle margin in
Figure~\ref{fig:oracle} then behaves as Eq.~\ref{eq:oracle-gap}
requires: on that cell the observed margin is $7$\,pt, matching
$\mathbb{E}[h]$ to the percentage point, which confirms that the decomposition is measuring what we think it is.

\subsection{When Does SC Hurt?}

A harm event requires a wrong answer to win a plurality over a
correct one in five samples, which points to the shape of the
model's distribution over candidate answers, so we measured that
shape on the cached traces. Pooling both datasets, the five
samples yield on average $3.8$ distinct answers per prompt for
Qwen-FP16 and $3.0$ for Gemma-FP16, and the winner carries only
$42\%$ and $58\%$ of the five votes. The plurality is thin, at
most two votes of five, on $72\%$ of Qwen-FP16 and $47\%$ of
Gemma-FP16 prompts. Harm concentrates in that thin regime: across
the four pooled cells, $80$--$90\%$ of harm events were decided by
a winner holding no more than two votes, and on $42$--$67\%$ of
them the correct answer was among the five samples and lost the
vote anyway. Much of the harm is thus an aggregation failure rather
than a sampling failure, which is why drawing more traces is not the obvious solution.

Quantisation moves every one of these quantities in the wrong direction.
Under Q4 the mean number of distinct answers rises (Qwen $3.8$ to
$4.2$, Gemma $3.0$ to $3.5$), the winning vote share falls ($42\%$
to $35\%$, $58\%$ to $48\%$), and thin pluralities become more
common ($72\%$ to $84\%$, $47\%$ to $62\%$). Flattening the answer
distribution is precisely what weakens a plurality vote, and the
direction matches the numerical-computation and planning errors
that low-bit quantisation is reported to induce on math
reasoning~\cite{Li2025QuantMath}.
The outcomes follow the same pattern: on Qwen-Q4${\times}$MATH-500 the help rate
falls $2$\,pt relative to Qwen-FP16 while the harm rate rises
$2.5$\,pt, inverting $\Delta_\mathrm{sc}$. Four cells at one seed
cannot separate quantisation from what correlates with it, so we
read this as a consistent association rather than an established
mechanism, though it does rule out the harm being an artefact of one model or dataset. The two rates are simply both small and
of the same order here, and a plurality vote cannot distinguish that case from the large-model one.

\section{The TTC-Gate}
\label{sec:method}

\subsection{Problem Formulation}

We want a scalar score function $\sigma:\mathcal{X}\to[0,1]$ that
maps each prompt, given its greedy-pass statistics, to a
``run-SC'' propensity. For threshold $\tau\in[0,1]$ the gate
policy is
\begin{equation}
\pi_\tau(x) \;=\;
\begin{cases}
\text{accept SC@}5 & \text{if } \sigma(x)\ge\tau,\\
\text{accept greedy} & \text{if } \sigma(x)<\tau,
\end{cases}
\label{eq:policy}
\end{equation}
with $a_\pi(x)\in\{0,1\}$ denoting the policy's correctness. Its
compute is Eq.~\ref{eq:cost} at
$p_{\mathrm{invoke}}=\Pr[\sigma(x)\ge\tau]$, written $C(\tau)$,
and its \emph{accuracy retention} is
$R(\tau)=\mathbb{E}_\pi[a_\pi]/\alpha_\mathrm{sc}$. The goal is to
maximise the skip rate $\kappa(\tau)=\Pr[\sigma(x)<\tau]$
(equivalently, to minimise $C(\tau)$) subject to
$R(\tau)\ge R_0$.

\subsection{Gate Instances}

We instantiate three gates, ordered by model capacity:

\paragraph{\textsc{conf\_only}.}
A single hand-crafted feature: the mean greedy-pass token
margin confidence
\(
\mathrm{mc}(x) = \frac{1}{T}\sum_{t=1}^{T}
\big(p_t^{(1)} - p_t^{(2)}\big),
\)
where $p_t^{(1)},p_t^{(2)}$ are the two largest probabilities of
the softmax over the top-$5$ logits at greedy-step $t$. The gate
score is $\sigma_{\text{conf}}(x) = 1 - \mathrm{mc}(x)$, min--max
rescaled to $[0,1]$ over the cell's prompts, a monotone map that
only sets the scale on which $\tau$ is defined. Nothing is learned and
$\tau$ is the only parameter.

\paragraph{\textsc{logistic}.}
A 12-feature $L_2$-regularised logistic regression. The feature
vector $\phi(x)\in\mathbb{R}^{12}$ concatenates four greedy-pass
statistics (mean margin confidence $\mathrm{mc}(x)$, mean top-$5$
entropy, mean full-vocab entropy, greedy token count) with eight
prompt-text counts (characters, words, digits,
arithmetic operators, question marks, commas, lines, and
multi-step connectives such as ``then'' or ``given''). Writing
$\tilde\phi(x)=\Sigma^{-1/2}(\phi(x)-\mu)$ for the train-fold
standardisation, the gate score is
\begin{equation}
\sigma_{\text{log}}(x) \;=\;
\big(1+e^{-(w^\top\tilde\phi(x)+b)}\big)^{-1},
\label{eq:logistic}
\end{equation}
fitted by class-weighted cross-entropy with $L_2$ penalty
$\lambda{=}1$, on the training label $y_i=y(x_i)$, i.e.\ ``SC
flips a wrong greedy answer to right''. Class weights
$\omega_{y_i}\!\propto\!1/\Pr(y=y_i)$ counter the label imbalance, and
scikit-learn defaults are used otherwise.

\paragraph{\textsc{mlp}.}
A two-hidden-layer MLP with widths $32\times 16$ over the same
$\tilde\phi(x)$, ReLU activations, $L_2$ regularisation on all
weights, trained on the same labels by unweighted cross-entropy,
as scikit-learn's MLP supports no class weights.

\subsection{Training and Threshold Selection}

Within each cell we run $5$-fold cross-validation over the $n$
prompts. Per fold, the standardisation $(\mu,\Sigma)$ and the
classifier are fit on the four training folds and applied to the
held-out fold, so each prompt is scored by a model that never saw
it. Pooling the five held-out folds gives one out-of-fold score per
prompt, $\{\sigma(x_i)\}_{i=1}^n$, on which the Pareto curves and
operating points of \S\ref{sec:pooled} are computed. The operating
threshold is
\begin{equation}
\tau^{\star} \;=\;
\arg\max_{\tau\in[0,1]}\; \kappa(\tau)
\quad\text{s.t.}\quad
R(\tau)\ge R_0,
\label{eq:tau}
\end{equation}
with $R_0=0.98$, asking for the largest compute saving compatible
with near-lossless accuracy. We then fix $\tau^{\star}$ and
bootstrap-resample prompts $1\,000$ times for $95\%$ CIs on
$(\kappa,C,R)$.

One consequence of this protocol should be stated before the
results. The scores are out of fold, but $\tau^{\star}$ is chosen
on the same out-of-fold pool on which we then report
$(\kappa,C,R)$, with no outer split around the selection, so the
reported retention is mildly optimistic by an amount $n{=}400$
cannot measure without a separate threshold set. Eq.~\ref{eq:tau}
also places $\tau^{\star}$ exactly where $R(\tau)$ has just
crossed $R_0$, the least favourable point for a reused estimate.

\subsection{Pooled-Cell Operating Points}
\label{sec:pooled}

\begin{figure}[tb]
\centering
\includegraphics[width=0.70\linewidth]{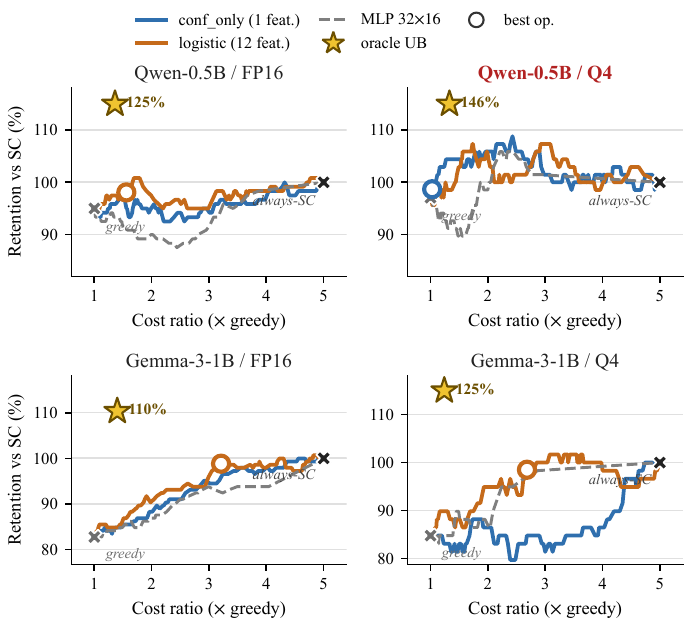}
\caption{TTC-gate Pareto curves on the four pooled cells
($n{=}400$ per cell). Each curve traces $\tau$ across the
$(C,R)$ plane. Gold star: oracle upper bound (annotated value
is the true retention, plot is clipped at 115\% for
readability). Hollow circle: best operating point at
$R\ge 0.98$. Italic labels mark always-greedy ($C{=}1$) and
always-SC ($C{=}5$). \textsc{Logistic} beats \textsc{conf\_only}
on 3/4 pooled cells, and \textsc{mlp} is best on none.}
\label{fig:pareto}
\end{figure}

Figure~\ref{fig:pareto} traces each gate's Pareto curve through
the $(C,R)$ plane on the four pooled cells ($n{=}400$), with gold
stars at the oracle and hollow circles at the best $\tau^{\star}$
under $R\ge 0.98$. On Qwen-Q4 the gate collapses to always-greedy
($C\approx 1.03$), the right response when SC adds only $0.5$\,pt over greedy; on the
other cells it lies at $C\approx 1.6$--$3.2$. \textsc{Logistic}
wins 3/4 pooled cells and \textsc{conf\_only} the fourth, with
\textsc{mlp} best on none.

\begin{table}[tb]
\centering\small
\caption{Pooled ($n{=}400$) TTC-gate operating points.
$\tau^{\star}$ is chosen by Eq.~\ref{eq:tau} with $R_0=0.98$.
The best gate skips $45$--$99\%$ of SC calls, which removes
$36$--$79\%$ of always-SC compute under Eq.~\ref{eq:cost}.
Bootstrap ($1\,000{\times}$) $95\%$ CIs in brackets. The skip-rate
CIs are tight, with half-widths under $5$\,pt. The retention CIs are not tight, with
lower bounds of $76$--$86\%$, because retention divides by an
$\alpha_\mathrm{sc}$ of $0.15$--$0.36$ on these cells. The
$98.1$--$98.8\%$ retention figures are therefore point estimates,
not intervals, and $\tau^{\star}$ was selected on this same
out-of-fold pool (\S\ref{sec:method}).}
\label{tab:gate-pooled}
\setlength{\tabcolsep}{2.3pt}
\begin{tabular}{@{}lcccccc@{}}
\toprule
Cell & Method & $\tau^{\star}$ & skip$\%$ $[$CI$]$ & $C$ $[$CI$]$ & $R$ $[$CI$]$ \\
\midrule
Qwen-FP16  & \textsc{logistic} & $0.625$ & $86.0\;[82.5, 89.3]$ & $1.56\;[1.43, 1.70]$ & $98.1\;[82.5, 112.5]$ \\
Qwen-Q4    & \textsc{conf\_only} & $0.915$ & $99.3\;[98.5, 100.0]$ & $1.03\;[1.00, 1.06]$ & $98.7\;[76.5, 120.6]$ \\
Gemma-FP16 & \textsc{logistic} & $0.46$  & $44.7\;[40.0, 49.5]$ & $3.21\;[3.02, 3.40]$ & $98.8\;[86.2, 112.4]$ \\
Gemma-Q4   & \textsc{logistic} & $0.515$ & $57.9\;[53.2, 62.7]$ & $2.68\;[2.49, 2.87]$ & $98.5\;[77.9, 122.0]$ \\
\bottomrule
\end{tabular}
\end{table}

Table~\ref{tab:gate-pooled} reports the numerical operating
points. On Qwen-Q4, \textsc{conf\_only} skips $99.3\%$ at
$C{=}1.03$: pooled SC accuracy is only $0.5$\,pt above greedy, so
the gate essentially disables SC, driven by the net-negative
Qwen-Q4${\times}$MATH-500 cell. Skip rates across the four cells
are $44.7$--$99.3\%$ with CI half-widths under $5$\,pt, removing
$36$--$79\%$ of always-SC compute. Retention is $98.1$--$98.8\%$ as
a point estimate but carries a bootstrap lower bound of
$76$--$86\%$, so we do not state the compute and accuracy claims as a single claim.

\subsection{Feature and Capacity Ablation}
\label{sec:ablation}

We compare \textsc{conf\_only} ($1$ feature), \textsc{logistic}
($12$ features) and \textsc{mlp} ($12$ features, $32\times 16$
hidden) at both the $n{=}200$ per-task scale and the $n{=}400$
pooled scale.

At the $n{=}200$ per-task scale the training signal is sparse: each fold trains
on $160$ prompts whose positive rate is only $4$--$12.5\%$, i.e.\
$6$--$20$ positive examples. At that size \textsc{mlp} collapses to
$\tau{=}0$ (``always run SC'') on six of the eight per-task cells,
and between the two usable gates the split is clear, with
\textsc{conf\_only} winning three of the four GSM8K cells and
\textsc{logistic} all four MATH-500 cells: one confidence feature
suffices where prompts are homogeneous, and the extra features help where they are not. Pooling to $n{=}400$ roughly
doubles the positives per fold, to $19$--$32$, and
\textsc{logistic} then overtakes \textsc{conf\_only} on $3/4$
cells.

The \textsc{mlp} is the informative negative result. It never wins at the
pooled scale, and across all twelve experiments it is the best gate
exactly once, on Qwen-FP16${\times}$GSM8K, where its $82\%$ skip
against $78\%$ for \textsc{conf\_only} lies inside the bootstrap CIs of both. Elsewhere it collapses to always-SC or falls behind a linear gate. The decision surface is evidently close to linear in
the scaled features, so the extra capacity can only fit fold-specific noise.

\begin{figure}[tb]
\centering
\includegraphics[width=0.67\linewidth]{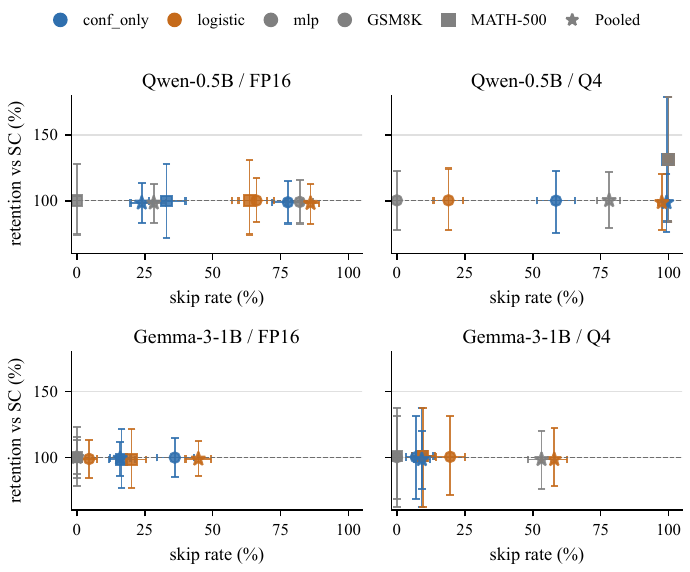}
\caption{Operating points in the (skip rate, retention) plane with
$1\,000{\times}$ bootstrap $95\%$ CI bars on both axes, one panel
per model-precision cell. Shape marks GSM8K (circles), MATH-500
(squares) and pooled (stars) evaluations, colour marks the gate,
and the dashed line marks always-SC parity. Pooled points carry
the tightest intervals; the per-task ones overlap between
methods.}
\label{fig:operating}
\end{figure}

In the pooled Qwen-FP16 \textsc{logistic} gate the five largest
standardised coefficients are four correlated prompt-length and
layout counts (commas, lines, words, characters,
$|w_j|=0.84$--$0.91$) plus greedy full-vocab entropy ($+0.81$).
Collinearity splits the length features' joint signal across
opposite signs, so the informative quantity is their weighted
combination and not any single coefficient. A surface this close to linear is consistent with the \textsc{mlp}'s failure to improve on it.
Figure~\ref{fig:operating} shows why we do not rank the gates
at $n{=}200$ at all: the per-task intervals overlap heavily
between methods, and only the pooled points are clearly separated.

\section{Discussion}

\paragraph{Why is always-SC dominated?}
The guarantees available for sampling-based
TTC~\cite{Cordero2025Certified,Zhou2025RPC} assume mild
concentration around the correct mode, and \S\ref{sec:oracle} is a
measurement of that assumption failing. It also explains why
methods that only shrink
$N$~\cite{Aggarwal2023AdaptiveConsistency,Aghazadeh2025CGES,Kim2026ReASC}
cannot repair it: fewer samples change the variance of the vote,
not the fact that the vote is taken over a distribution whose mode
is wrong.

\paragraph{Gate capacity and data scale.}
The ablation of \S\ref{sec:ablation} shows that capacity is not the limiting factor at this data scale, not that nonlinearity is
useless in principle: what the pooled logistic gate gains over
\textsc{conf\_only} is a weighted combination of prompt-length
features and greedy entropy that the confidence margin alone
misses, and the \textsc{mlp}'s single win lies inside the bootstrap noise.

\paragraph{Why is Gemma's gate weaker than Qwen's?}
Across Qwen cells the best gate skips $59$--$100\%$ of prompts, and across Gemma cells only $10$--$58\%$. Part of that gap is not a
property of the gate. Gemma's answer distributions are the more
concentrated ones (\S\ref{sec:oracle}) and its pooled harm rate is
roughly half Qwen's ($3.8\%$ against $7.5$--$7.8\%$), so SC is more
often worth
running and there is less to skip. That explains the direction but not the size of the gap. Gemma-3's post-training
pipeline~\cite{Gemma3Report} and choices such as sliding-window
attention could each compress the per-prompt spread of the features
the gate reads, and establishing which distributions actually
differ would take a study we have not run.

\paragraph{Composition and deployment.}
Adaptive-$N$
methods~\cite{Aggarwal2023AdaptiveConsistency,Wan2024RASC} act
after the invocation decision and the gate acts before it, so the
savings multiply: a skipped prompt costs one pass instead of five,
while cutting $N$ from $5$ to $2$ saves at most $2.5\times$ on a
prompt that still runs. The gate is otherwise drop-in, with no
retraining, no verifier, no sampler change, and one scalar
$\tau^{\star}$ re-tunable from a deployment's own $(C,R)$ curve. We
should be precise about what $36$--$79\%$ means, though: it counts
generation passes removed (Eq.~\ref{eq:cost}), not latency or
energy, and how much of it a phone obtains depends on batching,
prefill reuse and memory pressure our accounting does not
model~\cite{Wang2025MNNLLM,Bsharat2025MobileMMLU,Cao2026PER}. The
gate comes before any cross-model cascade~\cite{Zheng2025CITER,Ong2024RouteLLM,Chen2024FrugalGPT},
which makes selective SC a plausible first stage in a multi-level
TTC stack rather than a substitute for one.

\paragraph{Broader implications.}
The efficient-TTC literature treats SC as a monotone gain whose
only open question is how many samples to draw, but on the models
we test the prior question of whether to sample at all is the one that changes accuracy. Two consequences follow, both scoped to this
regime. Benchmark tables reporting SC@$N$ accuracy without the
help/harm split can conceal a set of prompts on which SC degrades
an answer greedy already had. And Eq.~\ref{eq:oracle-gap} is not
specific to plurality voting: any aggregation step that can turn a
correct base answer wrong has an oracle margin of exactly its own
harm rate, so the same check is worth running for best-of-$N$
reranking and self-refinement.

\section{Limitations}
\label{sec:limitations}

Our grid is two model families, both ${\le}1$B parameters, at two
precisions, on two math benchmarks, at $N{=}5$, with one
generation seed per cell. Code generation, commonsense question
answering, open-ended tasks, larger models, and reasoning-distilled
or reinforcement-learning-trained small models are untested. We would
expect distilled reasoners to concentrate the correct mode and so
shrink $\mathbb{E}[h]$, weakening the case for a gate, and larger
$N$ to shrink it as well, but neither expectation is measured
here. What we take to be transferable beyond the grid is the
decomposition and the practice of reporting a harm rate, not the
numeric ranges.

The statistical caveats add up. Every CI resamples prompts with
each prompt's greedy and SC outcomes held fixed, so it describes
sensitivity to the prompt sample and says nothing about run-to-run
sampling variance. The cell carrying the most weight in our
framing, the negative $\Delta_\mathrm{sc}$ on
Qwen-Q4${\times}$MATH-500, rests on $n{=}200$ prompts at one seed,
and two or three seeds on the two Qwen cells is the first
experiment we would add. The $98.1$--$98.8\%$ retention figures are
also point estimates whose CIs reach down to $76$--$86\%$
(Table~\ref{tab:gate-pooled}), since retention divides by an SC
accuracy of only $15$--$36\%$. Skip rate and compute are the well-supported claims, with CI half-widths under $5$\,pt. Retention is less well supported, especially because $\tau^{\star}$ is selected on the pool it is then
evaluated on (\S\ref{sec:method}).

Two further gaps are worth noting. Costs are counts of generation
passes under Eq.~\ref{eq:cost}, with the greedy trace treated as
one of the $N$ votes. We report no wall-clock latency, energy, or
peak-memory measurement, and an implementation unable to reuse the
greedy trace would have $25$--$79\%$ rather than $36$--$79\%$ of
always-SC compute removed. And while the gate is orthogonal to
adaptive-$N$ methods by construction, we neither implement
Adaptive-Consistency~\cite{Aggarwal2023AdaptiveConsistency},
confidence-guided stopping~\cite{Aghazadeh2025CGES,Kim2026ReASC},
budget prediction~\cite{Ji2025SeerSC} or cross-prompt
allocation~\cite{Liu2026PETS} as baselines, nor measure the gate
composed with them. A deployment choosing among these options
needs exactly that comparison, and its absence is the clearest gap
here. An adaptive $\tau$ and a three-way greedy / SC /
SC-plus-verifier gate are likewise untested.

\section{Conclusion}
\label{sec:conclusion}

For prompted CoT with SC on the small models we study, ``always
run SC'' is dominated by a per-prompt oracle gate in every one of
our $12$ experiments. The oracle's margin equals the rate at which
SC flips correct greedy answers to wrong, and at this scale that
rate is large enough to matter: $3$--$10.5\%$ of prompts, and on
one cell enough to leave SC worse than greedy overall. A gate on
greedy-pass signals recovers much of the available saving,
skipping $45$--$99\%$ of SC calls and removing $36$--$79\%$ of
always-SC compute, with retention point estimates of
$98.1$--$98.8\%$ whose intervals are wide enough that we report the
compute and accuracy claims separately. The gate is model-local,
verifier-free, and cheap enough to run on the device that hosts
the model. None of this resolves the question for larger models,
larger $N$, or non-mathematical tasks, but it does suggest a cheap practice: before reporting that SC@$N$ improves accuracy on a small
model, measure the harm rate and report it next to the gain.

\begin{credits}
\subsubsection{\ackname} This work was supported by the UK
Engineering and Physical Sciences Research Council (EPSRC) under
grant EP/X03688X/1. This work was also supported by the EPSRC
Centre for Doctoral Training in Distributed Algorithms. The
authors are grateful for the support of Stephen Pasteris at The
Alan Turing Institute.

\subsubsection{\discintname} The authors have no competing
interests to declare that are relevant to the content of this
article.
\end{credits}

\bibliographystyle{splncs04}
\bibliography{custom}

\end{document}